\documentclass[num-refs]{wiley-article}

\makeatletter
\newif\ifanonymous
\if@blindreview\anonymoustrue\else\anonymousfalse\fi
\makeatother

\usepackage{siunitx}
\usepackage{todonotes}
\usepackage{hyperref}
\usepackage{cleveref,xcolor,stex-logo,rustex}
\usepackage[inline]{enumitem}

\newlist{inlineenum}{enumerate*}{1}
\setlist[inlineenum]{}

\definecolor{lbcolor}{rgb}{0.9,0.9,0.9}  
\usepackage{listings}
\newcommand{\ALeA}{AL{\footnotesize E}A{ }}

\papertype{Original Article}
\paperfield{Journal Section}

\title{Leveraging Large Language Models to Generate Course-specific Semantically Annotated Learning Objects}

\author[1\authfn{2}]{Dominic Lohr}
\author[1\authfn{2}]{Marc Berges}
\author[3]{Abhishek Chugh}
\author[2\authfn{2}]{Michael Kohlhase}
\author[2\authfn{2}]{Dennis Müller}

\affil[1]{Professorship for Computer Science Education}
\affil[2]{Professorship for Knowledge Representation and Management}
\affil[3]{\url{sophize.org}}

\corraddress{Dominic Lohr, Professorship for Computer Science Education, Friedrich-Alexander-Universität Erlangen-Nürnberg, 91058, Germany}
\corremail{dominic.lohr@fau.de}

\presentadd[\authfn{2}]{Department of Computer Science, Martensst. 3, Friedrich-Alexander-Universität Erlangen-Nürnberg, Germany}

\fundinginfo{Federal Ministry of Education and Research Germany, Grant: 16DHBKI089}

\runningauthor{Lohr et al.}

\def\OMDoc{\textsc{OMDoc}}

\begin{document}

\begin{frontmatter}
\maketitle


\begin{abstract}
\textbf{Background}: Over the past few decades, the process and methodology of automated question generation (AQG) have undergone significant transformations. Recent progress in generative natural language models has opened up new potential in the generation of educational content.

\noindent \textbf{Objectives}: This paper explores the potential of large language models (LLMs) for generating computer science questions that are sufficiently annotated for automatic learner model updates, are fully situated in the context of a particular course, and address the cognitive dimension \texttt{understand}.

\noindent \textbf{Methods}: Unlike previous attempts that might use basic methods like ChatGPT, our approach involves more targeted strategies such as retrieval-augmented generation (RAG) to produce contextually relevant and pedagogically meaningful learning objects.

\noindent \textbf{Results and Conclusions}: Our results show that generating structural, semantic annotations works well. However, this success was not reflected in the case of relational annotations. The quality of the generated questions often did not meet educational standards, highlighting that although LLMs can contribute to the pool of learning materials, their current level of performance requires significant human intervention to refine and validate the generated content.

\end{abstract}

\end{frontmatter}





\vspace*{-3em}
\section{Introduction}
\label{sec:introduction}

The determinants of learning success have been extensively researched in various disciplines \cite{Hattie.2009, Kirby.2012}. Numerous empirical studies support the hypothesis that tailoring learning materials to learner's needs significantly increases the effectiveness of learning outcomes \cite{Sweller.1988, Sweller.1998}. The mastery learning theory by Bloom \cite{Bloom.1968} or the personalized system of instruction theory by Keller \cite{Keller.1968} postulate teaching methodologies that highly build on individualized learning materials and assessment on a very fine-grained level. However, resource limitations and an increasingly diverse educational landscape challenge teachers and content creators in general. 

Adaptive learning systems like 
\ifanonymous 
ANONYMISED FOR REVIEW
\else 
\ALeA \cite{kruseLearningALeATailored2023}
\fi
 promise to address this need by delivering Learning Objects (LOs), such as definitions,
examples, or questions tailored to individual learners' specific prior knowledge, competencies, and preferences. One way of realizing this is by using semantically annotated LOs that allow for determining their prerequisites, which concepts they address, and which competencies they intend to foster.
In conjunction with a model of a learner's current knowledge and competencies, this allows for many valuable services, such as the automatic generation of \emph{flashcards} or so-called \emph{guided tours}: individually selected sequences of LOs (``learning paths'') that can be automatically created by learners on demand by a simple click, to open up a specific concept in a specific cognitive dimension. To make this possible, the pool of semantically annotated LOs must be just as diverse as the educational biographies of the learners.
However, developing sufficiently annotated LOs requires huge amounts of manual labor and expertise.
This bottleneck underlines the need for innovative solutions that can generate customized learning materials automatically, thus becoming a focus of current research in computer science education. Attempts to (partially) automate the creation of quiz questions can be found in the literature as early as the 1970s \cite{wolfeAutomaticQuestionGeneration1976}. 

Studies on the role of artificial intelligence in education point out that a significant hurdle is the deficiency of suitable learning materials for individualized and adaptable learning \cite{dennyCanWeTrust2023}, a problem that may be mitigated by the capabilities of large language models (LLMs).
Research in computing education is exploring the capabilities of these models to generate educational content that is both contextually appropriate and educationally demanding, offering a promising solution to the constraints associated with manually creating content. However, the effectiveness of LLM-generated educational content is often limited by their lack of focus on specific course content and the learner's current understanding. The generated LOs often do not fit within the intended educational framework or meet the diverse needs of individual learners.

This paper presents the results of experiments on generating semantically annotated quiz questions using a state-of-the-art LLM. In particular, we investigate the question of to what extent LLMs can be used to generate questions in the domain of university-level computer science (CS) that are didactically valuable, are sufficiently annotated to allow for the above selection process, can ideally be graded by a software system (e.g. multiple choice or fill-in-the-blanks questions) to automatically and immediately increase the accuracy of the system's learner models, and are entirely situated in the context of a particular university-level course concerning terminology and notations used. Unlike earlier similar studies \cite{songAutomaticGenerationMultipleChoice2024, sarsaAutomaticGenerationProgramming2022}, our task requires extensive context, rendering it unsuitable for naive approaches, e.g., using ChatGPT. Instead, we use more targeted techniques beyond ``prompt engineering'', such as retrieval-augmented generation (RAG). 

\subparagraph{Overview} \Cref{sec:methodology} details our experimental methodology, including the considerations behind our choice of semantic annotations (\Cref{sec:stex}), the selection criteria for the LLM (\Cref{sec:model-selection}), the design of our question generation pipeline (\Cref{sec:pipeline}) and the evaluation framework used, including the criteria for assessing question quality and an expert-based survey methodology. The results of our experiments are presented and discussed in \Cref{sec:results}. Finally, \Cref{sec:conclusion} concludes the paper with a summary of our findings, implications for future research in automated question generation using LLMs, and potential pathways for enhancing the educational value of LLM-generated content in adaptive learning environments.


\section{Related Work}
\label{sec:related-work}

Over the past few decades, the process and methodology of automated question generation (AQG) have undergone significant transformations, driven primarily by advancements in computational linguistics and artificial intelligence~(AI). In the earlier stages, AQG relied heavily on rule-based systems that applied predefined templates and linguistic patterns to generate questions from text, requiring extensive manual crafting and domain-specific adjustments. First studies can be found by
~\citeauthor{wolfeAutomaticQuestionGeneration1976}~\cite{wolfeAutomaticQuestionGeneration1976}.

With the development of deep learning and neural network models, a transition to more sophisticated, context-aware systems can be recognized. These models, particularly sequence-to-sequence and transformer-based architectures, have enabled the generation of more nuanced, relevant, and diverse questions by ``understanding'' deeper semantic relationships within the text. \citeauthor{kurdiSystematicReviewAutomatic2020} conducted a systematic review of empirical research focused on addressing the issue of AQG within educational contexts \cite{kurdiSystematicReviewAutomatic2020}. They thoroughly outlined various methods of generation, tasks, and evaluation techniques found in studies between 2015 and early 2019. A standard method involves identifying sentences in the text sources with high information content using topic words or key phrases. The system then selects a keyword as the answer key, removes it to form a question stem, and generates incorrect choices (distractors) using a clustering method without external data. These systems mainly produce questions testing remember \emph{factual} knowledge, not \emph{understanding}, a key focus of the work presented in this paper, and the results show that large language models (LLMs) can potentially benefit from semantics-based approaches to generate meaningful questions that are closely related to the source content. 

Recent progress on generative natural language models has opened up new potentials in the generation of educational content \cite{pratherRobotsAreHere2023, shravyabhatAutomatedGenerationEvaluation2022,wangHumanLikeEducationalQuestion2022, wongHypeInsightExploring2024}. In recent years, more and more approaches have been found to generate tasks in a single step using LLMs like GPT-3 instead of dividing the AQG process into several sub-tasks.
\citeauthor{yanPracticalEthicalChallenges2024} did a systematic scoping review of articles published since 2017 to pinpoint the current state of research on using LLMs~\cite{yanPracticalEthicalChallenges2024}. They identified content generation, including multiple-choice questions and feedback generation, as primary educational tasks that research aims to automate. 
\citeauthor{mcnicholsExploringAutomatedDistractor2023} tested the effectiveness of LLM-based methods for automatic distractors and feedback generation using a real-world dataset of authentic student responses \cite{mcnicholsExploringAutomatedDistractor2023}. Their findings show that fine-tuning was ineffective and that other approaches than LLM-prompting need to be explored.
\citeauthor{dijkstraReadingComprehensionQuiz2022} \cite{dijkstraReadingComprehensionQuiz2022} developed a quiz generator based on OpenAI's GPT-3 model, fine-tuned on text-quiz pairs to generate complete multiple-choice questions, with correct answers and distractors. 
They noted that while most of the generated questions were of acceptable quality, creating high-quality distractors was more challenging than generating question-and-answer pairs. 

More specifically, LLM-based approaches have recently been applied in programming education (for an intensive literature review, see \cite{pratherRobotsAreHere2023}). \citeauthor{sarsaAutomaticGenerationProgramming2022} \cite{sarsaAutomaticGenerationProgramming2022} explored the capabilities of OpenAI's LLM \texttt{Codex} to generate programming exercises and code explanations. Their results show that the generated questions and explanations were novel, sensible, and, in some cases, ready to use. 
\citeauthor{tranGeneratingMultipleChoice2023} \cite{tranGeneratingMultipleChoice2023} evaluated the capabilities of OpenAI's GPT-4 and GPT-3 models to generate isomorphic multiple choice questions (MCQs) based on MCQ-stems from a question bank and an introductory computing course. Their findings underscore that the newer generation of LLMs outperforms older generative models in AQG.

\paragraph{Research Gap}

Reviewing the current landscape of research in AQG, several critical gaps emerge that necessitate further investigation. Current research shows that methods for generating closed-format tasks that target the cognitive dimension of \texttt{remember} \emph{factual} knowledge work well. However, generating tasks that target deeper understanding, such as those that require comprehension or analytical skills, remains a significant challenge. We also could not find any research on generating questions containing semantic annotations. 
While it is already known that LLMs have the potential for generating feedback, and in particular code explanations, no work attempts to generate questions using LLMs that go beyond introductory courses in computing education. \citeauthor{kurdiSystematicReviewAutomatic2020}\cite{kurdiSystematicReviewAutomatic2020} point out that there is little focus on controlling question parameters like difficulty and generating feedback. Since we want to incorporate the generated questions into an adaptive learning assistant, we have carefully considered which parameters are relevant for adaptive learning -- such as difficulty, cognitive dimension, and prior knowledge -- and want to explore how these can be integrated into the question generation process. We also did not find any studies that investigated generating questions using advanced LLM approaches such as RAG, as it is crucial for scalability beyond the original training data of the model.


\section{Requirements for the Generated Questions}
\label{sec:goals}

\vspace*{-1em}
We posit that automated question generation (AQG) in the context of university courses poses additional challenges barely (or not at all) covered by the existing literature, especially in a domain like \emph{computer science}: 

Firstly, we claim that in more abstract domains like mathematics and math-related subfields of computer science, questions solely focusing on remembering facts, or rudimentary application exercises are less suitable for the outcomes of a university-level class. This makes designing appropriate questions a more challenging task because doing so requires a level of understanding of and experience with the learning material. 

Secondly, although the topics covered in a particular course are common across different universities, the details are much less standardized (e.g., precise definitions, terminology, and notational conventions). This entails the additional requirement that questions (whether automatically generated or not) need to be formulated and situated in line with the conventions in a particular course, which are also more frequently subject to change. This makes approaches based on large amounts of dedicated training data, which necessarily do not generalize beyond a specific course \emph{and} instructor's preferences, unsuitable.
Instead, we should be able to provide the generator with the relevant learning materials (course notes, slides, etc.) whenever we want to generate a new batch of questions.

\vspace*{-1em}
\subparagraph{} Additionally, our goal is to utilize the generated questions in an adaptive learning assistant that is capable of automatically selecting suitable questions based on the associated \emph{learning objective} and a particular student's (estimated) prior knowledge in the form of a \emph{learner model}. Therefore, we need to annotate the questions with the relevant information for this selection process, namely: 
\begin{enumerate}
\item The \emph{concepts} the question is intended to test, 
\item the \emph{cognitive level} the question targets (modeled as levels in Bloom's revised taxonomy~\cite{andersonTaxonomyLearningTeaching2001}), and 
\item the \emph{prerequisite} concepts occurring in the question and the associated competencies that a user should have mastered for the question to be suitable. 
\end{enumerate} 
While this is less difficult for an experienced user, providing these annotations is still time-consuming and potentially automatable. Therefore, we also investigate the extent to which a large language model can directly generate fully annotated questions.

\begin{figure}
    \centering
    \includegraphics[scale=0.3]{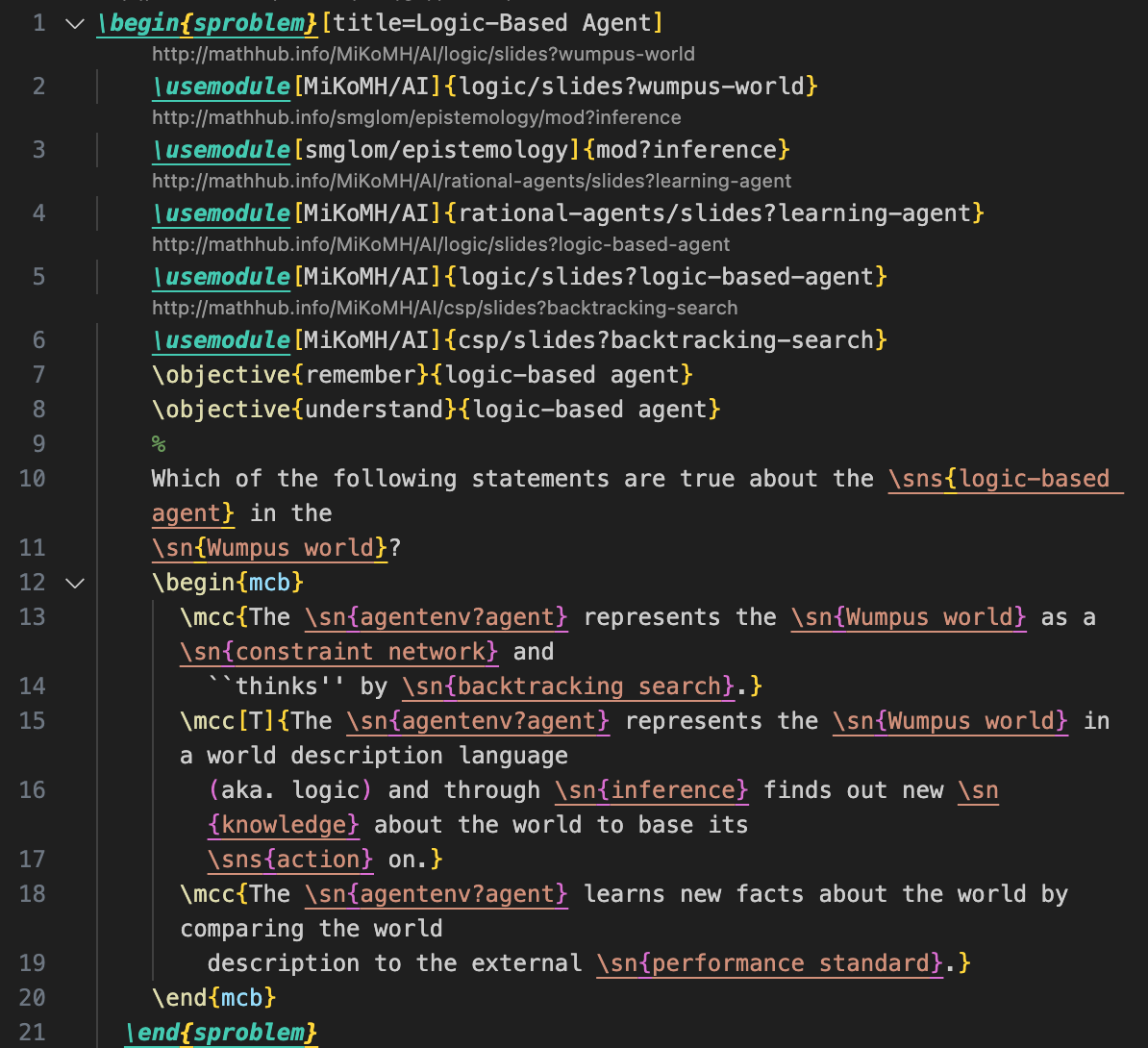}
    \includegraphics[scale=0.2]{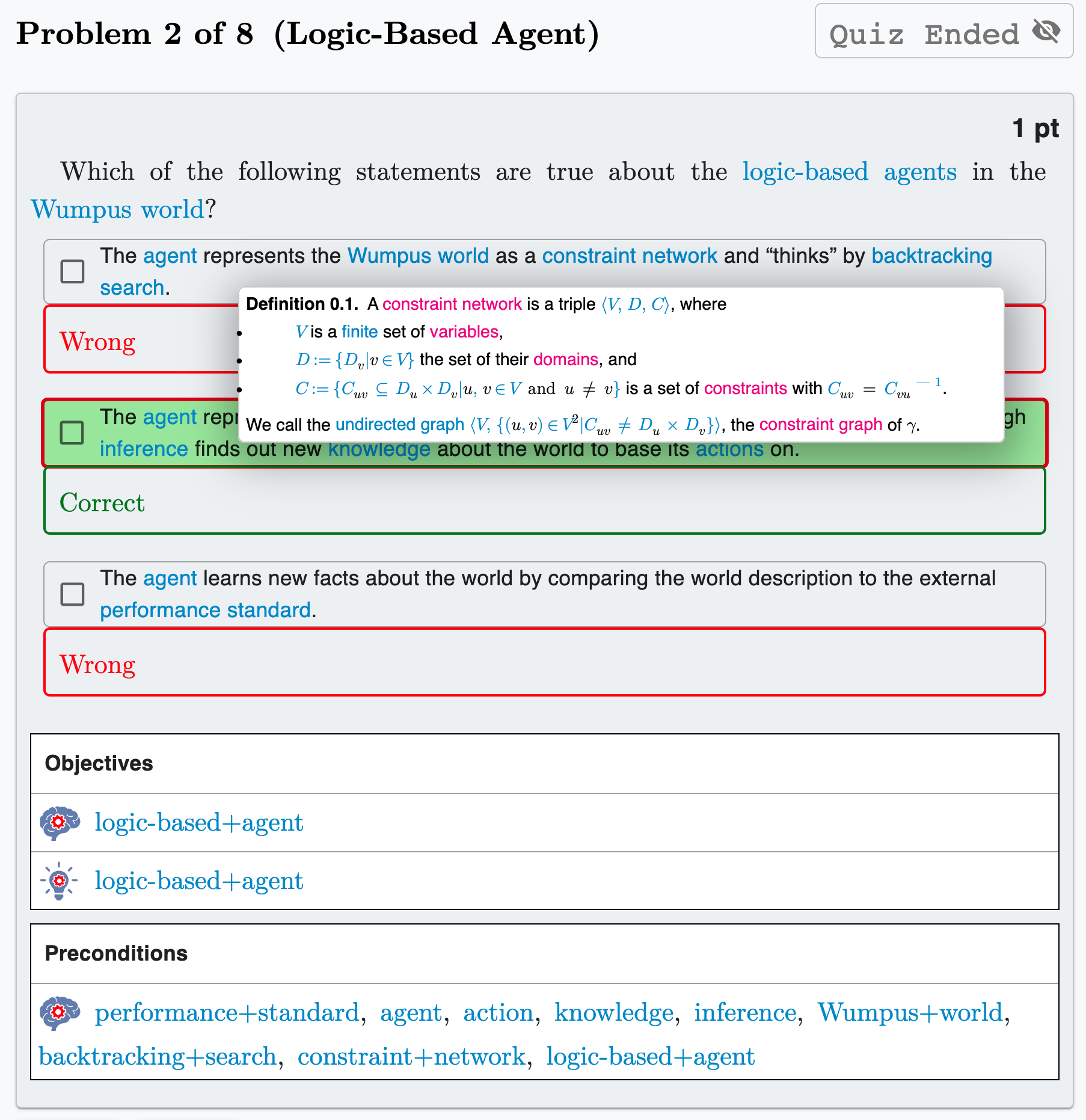}
    \caption{Source Code and Representation in the system}
    \label{fig:example}
\end{figure}

\Cref{fig:example} shows an example of a fully annotated multiple-choice question created by hand in \LaTeX{} following the Y-model framework \cite{lohrYModelFormalizationComputer2023}. The relational semantic annotations allow the learner to, e.g., hover over concepts in the text (after conversion to HTML) and display definitions (this functionality can, of course, be deactivated if desired). A detailed description of the specific annotation schema can be found in \Cref{sec:stex}.

Ensuring the quality of the generated tasks is a primary goal. As there is no standard metric in the literature for measuring the quality of questions \cite{chAutomaticMultipleChoice2020}, we reviewed metric lists from literature reviews on automatic question generation \cite{mullaAutomaticQuestionGeneration2023, chAutomaticMultipleChoice2020} to identify relevant criteria that determine the quality of the questions we generate, namely:
\vspace*{-1em}

\begin{itemize}
    \item correct technical language,
    \item appropriateness for a particular course context, including definitions, annotations, and conventions,
    \item feasibility of solving the question with the provided details and available course materials,
    \item clarity and lack of ambiguity,
    \item relevance to achieving the intended learning outcomes,
    \item alignment with the specific format of the task, and
    \item automatically gradable (closed format).
\end{itemize}

Additionally, feedback is one of the most influencing factors for learning success \cite{hattiePowerFeedback2007}, and recent experiments show the potential of large language models for generating feedback in programming education \cite{kieslerExploringPotentialLarge2023, lohrLetThemTry2024, hellasExploringResponsesLarge2023, balseInvestigatingPotentialGPT32023, matelskyLargeLanguageModelassisted2023}. A further goal is to generate questions containing feedback that helps learners understand why a given answer is wrong. 

The research questions are as follows:\\ (RQ1) To what extent can LLMs be used to generate university-suitable autogradable questions in CS education? \\
(RQ2) To what extent can LLMs be used to annotate these questions semantically?

\section{Methodology}
\label{sec:methodology}

\subsection{Semantic Annotations in \sTeX}
\label{sec:stex}
To semantically annotate learning objects -- including questions -- we use the \sTeX package~\cite{MueKo:sdstex22} for \LaTeX. \sTeX uses an ontology based on \OMDoc~\cite{Kohlhase:OMDoc1.2}: Concepts are represented as \emph{symbols} that can be introduced via the \texttt{\textbackslash symdecl}-macro and can be related to each other in various ways. Symbols are always declared in \emph{modules} (via the \texttt{smodule}-environment), which can \emph{import} (the symbols of) other modules, adding another layer of relations and allowing for sharing concepts among large collections of disparate documents, thus enabling the collaborative and modular development of domains of knowledge as highly interrelated knowledge graphs independent of, and across, \emph{presentational} context: symbols can have arbitrarily many definitions, formulations thereof, and different notations, to accommodate author's preferences without duplicating the domain knowledge itself.

Symbols can be \emph{referenced} in various ways; most importantly, the \texttt{\textbackslash symref}-macro allows annotating arbitrary text as representing a particular symbol, and formal notations can be produced via dedicated \emph{semantic macros} that additionally associate the notation with the corresponding symbol. Since the details are largely irrelevant to the topic of this paper, we refer to \cite{stexMan:on} for details and \Cref{sec:prompt-design} for examples.

More importantly, for our purposes, \sTeX also bundles the \texttt{problem}-package, providing dedicated markup macros and environments for various variants of quiz questions. Most importantly, it provides the \texttt{sproblem}-environment to annotate questions, within which we can use the \texttt{mcb} and \texttt{scb} environments for \emph{multiple} and \emph{single choice blocks}, respectively. As a third (autogradable) question type, the \texttt{\textbackslash fillinsol} macro can be used to generate a blank box for \emph{fill-in-the-blanks} questions. In all three cases, we can mark (or provide) the correct answer(s), add feedback text to be displayed if a particular answer (correct or wrong) is chosen, and specify \emph{grading actions} (i.e., set, add or deduct points) depending on the answers given. Additionally, we can specify the \emph{learning objectives} and \emph{prerequisites} as pairs of a \emph{symbol} and one of the keywords \texttt{remember}, \texttt{understand}, \texttt{apply}, \texttt{analyze}, \texttt{evaluate}, \texttt{create}. Symbols referenced in the body of the question are automatically determined to be prerequisites with the cognitive dimension \texttt{remember}. For example, if a module declares the symbol \texttt{plus} for addition on integers, then the question ``\texttt{what is 2 \textbackslash symref$\{$plus$\}\{$added to$\}$ 2?}'' automatically has the prerequisite (\texttt{remember},\texttt{plus}). We refer to such annotations as \textbf{relational annotations} and to those that do not relate text to some symbol (e.g. \texttt{fillinsol} or \texttt{sproblem}) as \textbf{structural annotations}.

Besides being \LaTeX{} compilable to \texttt{pdf}, we can use the \RusTeX system~\cite{Mue:rustextug23} to convert the \sTeX documents to HTML, preserving the semantic annotations in the form of attributes on the HTML nodes. Additionally, this assigns a globally unique URI to the document itself and every section, module, symbol, and learning object therein, enabling searching, querying, and referencing all of these afterward \emph{across} documents. 

Subsequently, we can embed various services that act on the semantic annotations directly into the HTML documents via JavaScript. Our learning assistant utilizes this to render quiz questions as interactive components, evaluate answers provided by students, display the appropriate feedback provided, and update a student's learner model accordingly (see \autoref{fig:example}).

While our usage of \sTeX is primarily motivated by our learning assistant, for the purposes of this paper, it offers additional advantages: Its underlying ontology, being designed around representing the semantics of mathematical statements formally, is consequently sufficiently expressive to subsume most semantic annotation systems, in particular those tailored specifically for quiz questions. Our results should, therefore, (all else being equal) generalize to other annotation schemes. Furthermore, since \sTeX inherits its syntax from \LaTeX, the LLM used does not need to be finetuned or explicitly prompted on the usage of a new or esoteric language -- the huge amount of publicly available \LaTeX~code implies that its syntax is well represented in the training data used for these models, and can therefore be expected not to pose an issue for generation. Indeed, our experiments did not yield any LLM output with basic syntax errors.

\vspace*{-1em}
\subsection{Model Selection}
\label{sec:model-selection}

Our goals put notable constraints on the methodology for generating questions: \begin{enumerate} \item Targeting \emph{understanding} rather than \emph{remembering} factual knowledge implies that the model we use should be capable of ``synthesizing'' complex knowledge, excluding, e.g., smaller language models less capable of ``reasoning''. \item Since the knowledge domain and preferred conventions should not be fixed, deliberately training a model on dedicated course materials is largely not feasible, which calls for \emph{few-shot} approaches. And \item since the latter point also entails providing possibly large amounts of course materials \emph{online} during generation, the model used needs to be able to process large amounts of text at once. More precisely, it should allow for a large enough \emph{context window} -- the maximum number of tokens the model can consider in a single prompt/reply step. \end{enumerate}

OpenAI's GPT-4~\cite{openaiGPT4TechnicalReport2023} shows consistently better results across all tasks compared to alternative large language models\footnote{See, e.g., \url{https://toloka.ai/blog/llm-leaderboard/} for a regularly updated comparison across several dimensions}. In particular, we initially experimented with the free GPT-3.5 model (that powers the free version of ChatGPT), which quickly demonstrated very poor performance both concerning the questions generated and, more broadly, the ability to follow the instructions provided -- meaning more prompt design was unlikely to improve the results significantly.

Similarly, most models (including almost all open-source models) primarily focus on short conversations and, therefore, only support a relatively small context window. For example, one of the currently most popular open source language models, \emph{LLaMA}~\cite{touvron2023llama}, can process a total input of 2048 tokens, whereas GPT-3.5 and GPT-4-Turbo have context windows of up to 16,385 and 128,000 tokens, respectively.\footnote{It should be mentioned, that the precise meaning of ``context window'' is not necessarily clear. There is some speculation that the huge context window advertised for GPT-4-Turbo involves neural compression and similar (potentially lossy) techniques to allow for more input rather than representing the actual size of the underlying transformer model's input layer.}

We consequently opted for the (unfortunately commercial, closed source, and proprietary) GTP-4-Turbo model as the one holding the most promise with respect to our criteria at the time of the experiments. Unfortunately, this also means we can not provide access to a public instance of our pipeline since the API used to access the model is monitored and billed on a per-token basis.\footnote{A common problem is people actively crawling the internet for API keys and publicly accessible APIs to GPT models, intending to abuse those for unintended purposes (e.g., via \emph{``prompt hacking''}).}

\vspace*{-1em}
\subsection{Overview of the Generation Pipeline}\label{sec:pipeline} To allow for the model to generate questions for a \emph{specific course} regarding terminology, definitions, and notations, we opted for a technique called \emph{retrieval-augmented generation (RAG)}. This technique gives models access to additional information beyond their training data by querying an external knowledge base (such as a database or a web search engine) for results relevant to the specific prompt. It concatenates them to the prompt itself before passing it into the model.

To explain how we use RAG, we give a brief overview of the entire question-generation pipeline. Note that we deliberately leave room for variation: \vspace*{-1em}
\begin{itemize}
    \item A course instructor selects some concept (represented by an \sTeX symbol), a cognitive dimension, a document of course materials (i.e., the entire lecture notes) via its URI, and additional parameters (e.g., number of questions, difficulty level, types of questions, a brief description of the course topics, ...).
    \item The system replaces placeholders in a generic \emph{``master prompt''} (see \Cref{sec:prompt-design}) by the parameters provided.
    \item The system then selects those text fragments in the provided course materials that directly relate to the chosen concept -- e.g., definitions, examples, surrounding remarks, etc. -- on the basis of the semantic annotations in the document and appends them to the master prompt. Since \sTeX symbols are intrinsically relational, we can, in principle, also add materials for \emph{dependent concepts} down to some cutoff point. In practice, we currently pick the entire chapter (or section or subsection) in which the concept is introduced.
    \item The final prompt is passed to the LLM and its output is presented to the instructor.
\end{itemize}

We focus here on the scenario where all parameters -- in particular, the concept, cognitive dimension, and number of questions -- are explicitly fixed in the prompt, as this allows for better evaluation of the results of our experiments. However, we note that we can easily generalize to broader applications by automatically selecting some of the parameters. For example, we can generate questions for a whole \emph{chapter} along \emph{all} cognitive dimensions by determining the concepts introduced therein (which we can do automatically based on semantic annotations) and repeatedly running the above pipeline with, e.g., randomly varied parameters.

\subparagraph{} We also chose not to split the task into multiple prompts for specific purposes. For example, we could conceptually use three distinct prompts (or even distinct \emph{models}) for 1. generating questions, 2. adding feedback for students, and 3. introducing semantic annotations. This split would be natural in a non-LLM-based approach since all three steps require distinct methodologies and tools. However, when using the same model for all three steps, it is less clear whether splitting the task is advantageous. We leave the question of whether it is for future work but note that our initial (and admittedly superficial) experiments in that direction did not yield noticeably different results. Furthermore, experiments with instructing the model to (paraphrased) \emph{first} ``think of a good question, summarize it, and \emph{then} state the question as fully annotated \sTeX'' -- conceptually splitting the task into two separate steps -- did not seem to make a clear difference in the quality of the output either. This is to some extent unexpected since it closely corresponds to the popular \emph{``think step by step''}-instruction that is considered to generally improve results in LLM prompts related to reasoning.

\vspace*{-1em}
\subsection{Prompt Design}
\label{sec:prompt-design}

\vspace*{-0.5em}
Despite ongoing attempts to declare ``prompt engineering'' a marketable skill, few hard principles for designing LLM prompts can be consistently well supported empirically. The most important rules of thumb can be briefly summarized thusly~\cite{Zamfiresu-Pereira.2023,openaipromptengineering}: \begin{enumerate} 
    \item Be \emph{specific} in what the output should look like, 
    \item be \emph{detailed} by including any and all information and requirements relevant to the expected output, 
    \item provide \emph{examples} in the prompt, effectively transforming the task from a \emph{zero-shot} to a \emph{few-shot} approach, and, most importantly, 
    \item \emph{iterate} by repeatedly testing, observing problems with the outputs, and modifying the prompt to discourage these problems in subsequent iterations. 
\end{enumerate} 
Paraphrasing~\citeauthor{Zamfiresu-Pereira.2023}~\cite{Zamfiresu-Pereira.2023}, it helps thinking of the language model as an \emph{interpreter} and prompting as analogous to \emph{programming} in the sense that it requires clear and detailed instructions to an entity keen to ``do what you \emph{say}, not what you (clearly) \emph{mean}''.
 
Consequently, our final prompt is in many places the result of flaws occurring in previous iterations, which we will note where appropriate.

\subparagraph{} Our prompt starts with a brief summary of what we expect the model to do:
\begin{lstlisting}
Your task is to generate quiz questions that evaluate the competency of 
university students with respect to a number of particular concepts in a 
particular university course. You will be given 
- the names of the concepts, 
- the name of the course,
- a cognitive dimension to test for, 
- ...
\end{lstlisting}
We then clarify what we mean by \emph{cognitive dimension}. Note that the huge amounts of data the model was trained on also contain many texts about concepts in education; particularly Bloom's revised taxonomy~\cite{andersonTaxonomyLearningTeaching2001}. Nevertheless, we prefer to be precise:
\begin{lstlisting}
We use the term "cognitive dimension" according to Bloom's taxonomy. It is 
provided as one of the strings "remember", "understand", "apply", "analyse", 
"evaluate", or "create". Here is a short explanation for the taxonomy, 
together with verbs that are commonly used in questions that fit this taxonomy 
level:
- remember: recall facts and basic concepts (verbs: define, duplicate, list, 
  memorize, repeat, state)
- understand: ...
\end{lstlisting}

Note also, that we could save on tokens by restricting the prompt to only the one cognitive dimension that a user actually selects in the pipeline. In practice, the instructional part of the prompt is only a small part of the final prompt; the bulk of which consists of the learning materials provided. As such, there is little need to be conservative here, and listing all dimensions might help with clearly delineating them.

Next, we give a brief explanation of the \sTeX syntax used by the learning objects provided as context:
\begin{lstlisting} 
The learning objects will be given as a list of LaTeX snippets, using the sTeX 
package for semantic markup. In sTeX, concepts are declared using the \symdecl 
or \symdef commands, and can be referred to using...
\end{lstlisting}
Indeed, we can keep this part brief since the learning objects will naturally contain large amounts of \sTeX syntax that (ideally) allow the model to pick up on the relevant macros and their usage. In particular, we can safely omit example usages here.

We also make sure that sufficient information is provided in the prompt to ensure the reuse of symbols present in the learning objects and the surrounding modules:
\begin{lstlisting}
Every learning object will be prefixed with an id that indicates which file it 
is from, in a format close to what is required in \usemodule ...
\end{lstlisting}

Next, we explain precisely what kinds of questions we expect from the model, namely one of the three autogradable question types supported by \sTeX (\emph{multiple choice}, \emph{single choice}, or \emph{fill-in-the-blanks}). We then provide examples for each of them, so the model ``knows'' to replicate the relevant \sTeX macros and environments. We make sure that the examples are representative of the output we expect from the model, i.e., they should contain as many detailed annotations as possible, be (in our estimation) interesting and didactically valuable, and contain good feedback, especially for the wrong answer options. Therefore, we also point out that adding feedback is desired:

\begin{lstlisting}
A quiz question can be either a multiple choice question, a single choice 
question or a fill-in-the-blanks question. We can also provide feedback for 
each answer option using the `feedback` key of the relevant macros. This 
feedback is shown to the student after they have submitted their answer.

An example for a multiple choice question is the following; note the semantic 
markup:
```
\begin{sproblem}
  \usemodule[smglom/sets]{mod?bijective}
  \usemodule[smglom/sets]{mod?relation-composition}
  \usemodule[smglom/arithmetics]{mod?natarith}
  \objective{understand}{bijective}
  \objective{understand}{injective}
  \objective{understand}{surjective}

  Assume $\fun{f,g}\NaturalNumbers\NaturalNumbers$.  Which of the following are 
  true?
  \begin{mcb}
    \mcc[F,feedback={No, $f$ and $g$ are unrelated}]
      {If $f$ is \sn{injective}, so is $g$.}
    \mcc[F,feedback={No. since $f$ does not need to be \sn{surjective}, the 
      \sr{surjective}{surjectivity} of $g$ is not enough to make the 
      \sr{compose}{composition} of $f$ and $g$ \sn{surjective}.}]
      {If $f$ is \sn{injective} and $g$ is \sn{surjective}, then 
        $\compose{g,f}$ is \sn{surjective}.}
    \mcc[F,feedback={No. Since $f$ need not be \sn{surjective}, the 
      \sn{composition} need not be \sn{surjective} either.}]
      {If $f$ is \sn{injective} and $g$ is \sn{surjective}, then 
      $\compose{g, f}$ is \sn{bijective}.}
    \mcc[T]{If $f$ and $g$ are \sn{injective}, so is $\compose{g,f}$.}
    \mcc[T]{If $f$ and $g$ are \sn{surjective}, so is $\compose{g,f}$.}
  \end{mcb}
\end{sproblem}
```
The question is answered correctly, if the student selects exactly the \mcc 
options marked with [T].
\end{lstlisting}

We delimit the \sTeX code example using the three backticks \lstinline|```|, as is standard in, e.g., Markdown. In particular, it is the same delimitation used by GPT (in particular ChatGPT) when producing code in its output.

Furthermore, we deliberately chose an example of a \emph{multiple choice} question with more than one correct answer -- like (in our experience) human beings, GPT too seems to prefer having a single correct choice; in both cases likely because it is easier to come up with questions where there is one correct answer. To direct the model to generate more question types than just ``single choice questions disguised as multiple choice'', we promote a variety of question formats.

We subsequently provide similar examples for \emph{single choice} and \emph{fill-in-the-blank} questions. Finally, we reiterate the expected outcomes and add specific criteria we want the questions to satisfy, starting with ones that seemed important to us from the start:

\begin{lstlisting}
Once you are given the data described above, you are to reply with the given 
number of quiz questions as LaTeX code with semantic markup.

Make sure that the questions you generate satisfy the following criteria:
- The answer to a question should give a good indication regarding the extent 
  to which the student has mastered the concept with respect to the given 
  cognitive dimension.
- The questions should be as diverse as possible, i.e. they should test for 
  different aspects of the concept.
- Make sure you provide good feedback, especially for wrong answers, so that 
  students can learn what they did wrong.
- Assume the students know nothing about the concept other than what's in the 
  learning objects provided.
- In single/multiple choice questions, make sure that the distractors are 
  superficially plausible to students who have not yet mastered the subject, 
  as to not make the problems too easy.
- Feel free to use examples the students might know from elsewhere, e.g. basic
  high school level knowledge, or foundational principles of the study program 
  the course is in.
\end{lstlisting}

Finally, we list the parameters provided by the user as replacement variables that are substituted by the system:

\begin{lstlisting}
%%%%
concepts: %%CONCEPT%%
course: %%COURSE%%
...
learning objects: %%LEARNING_OBJECTS%%
\end{lstlisting}

\paragraph{Notable updates during iteration}
While some aspects of the prompt discussed so far have changed during iterating, the ``basic outline'' of the prompt seemed to work surprisingly well and has thus remained largely stable. In response to mistakes we noticed, we added additional criteria to the list at the end, which we discuss now.

\subparagraph{} Occasionally, there was an excessive tendency in the output to refer to the course materials directly, resulting in generated questions about e.g., specific names of variables in some example somewhere in the course notes. Hence, we attempt to avoid questions in the output about irrelevant specifics of the course materials:
\begin{lstlisting}
- We do not want students to rote-memorize definitions, examples, or other text
  in the learning objects, so never make the correct answer dependent on such
  details (e.g. variable names, particular examples, etc.).
\end{lstlisting}

It also sometimes added feedback or explanations \emph{to the text of the answer choice}, inevitably revealing the correct and wrong answers in the questions itself:
\begin{lstlisting}
- Do not put any text in the LaTeX code that directly states which answer is 
  correct - the sTeX macros used above take optional arguments explicitly for 
  that purpose.
\end{lstlisting}

In some instances, quiz questions were generated to ask for additional \emph{free-text} answers, so we emphasize that students will not be able to do so:
\begin{lstlisting}
- Note that students are limited to replying to a question in the form the 
  question type is posed in, i.e. ticking boxes in single/multiple choice 
  questions, or filling in a short text in fill-in-the-blanks questions. 
  They can not provide any additional text.
\end{lstlisting}

The output rarely contained fill-in-the-blank questions. When we modify the prompt to explicitly encourage those, problems quickly emerge, which we attempted to fix thusly:
\begin{lstlisting}
- The correct answer must be unambiguous, particularly for 
  fill-in-the-blanks questions.
- Importantly, the evaluation of student's answers in \fillinsol is done 
  automatically via string matching, so \fillinsol can only contain plain 
  text, no LaTeX code, and students need to type in the answer exactly to 
  get any points.
\end{lstlisting}
This seemed to help somewhat, but the primary noticeable result was even fewer fill-in-the-blank questions.

This should not be entirely surprising: beyond the \texttt{remember} dimension, fill-in-the-blank questions (where answers are literally string matched to a reference solution) are primarily helpful to ask for an answer that (ideally) can only result from, e.g., correctly applying an algorithm, computing some expression, counting specific properties, etc. -- i.e., they are very useful to make questions asking for \emph{complex reasoning} and \emph{processes} autogradable. These questions tend to be particularly difficult to design, though. Hence, the model strongly prefers multiple and single-choice questions.

In general, the generated questions for the \texttt{apply} dimension were not satisfactory. One strategy we tried to improve this was to explicitly point out the use of fill-in-the-blank questions in the prompt, but this did not lead to any noticeable improvement. 

\newpage

\subsection{Evaluation}
\label{sec:evaluation}
In the literature, there are two established techniques to evaluate the quality of generated questions \cite{mullaAutomaticQuestionGeneration2023}: (1) automatic evaluation and (2) human-based evaluation. 

Certain criteria outlined in Section \ref{sec:goals} necessitate specific insights from the course context, accessible only to individuals within the lecture environment. Similarly, assessing the quality of \sTeX annotations requires expertise and cannot be effectively automated. Hence, an expert-based evaluation was selected -- the most common evaluation approach in the context of automatic question generation \cite{mullaAutomaticQuestionGeneration2023}.

For this purpose, we (1) developed a structured survey to evaluate the quality of generated questions based on the criteria above, (2) generated a collection of 30 questions, (3) conducted the survey with experts within the course environment, and (4) conducted a qualitative analysis of the survey findings.

\subsubsection{Survey Design}

The survey's introductory segment supplied experts with context by outlining the parameters used in the question generation prompt, followed by the complete presentation of the generated question (GQ), incorporating any semantic annotations.

A free-text response field was incorporated to evaluate the accuracy of the content and detect any potential errors in the GQ. This section allowed experts to pinpoint specific content discrepancies or inaccuracies within the GQ. 
Additionally, we employed a 5-point Likert scale, spanning from ``Very Difficult'' to ``Very Easy,'' to assess the perceived level of difficulty of the GQ.

The survey consisted of six statements to assess the generated question's different quality dimensions. Experts were tasked with indicating their degree of agreement with these statements on a 7-point Likert scale, ranging from ``Strongly Disagree'' to ``Strongly Agree''. The statements were outlined as follows:

\begin{itemize} 
    \item The GQ has a good FIT in terms of teaching material.
    \item The GQ can be solved using the available teaching material.
    \item The task description of the GQ cannot be misinterpreted (is not ambiguous).
    \item The GQ is relevant for the achievement of the specified Learning Objective.
    \item The feedback provided for the answer options of the GQ is helpful.
    \item The structure of the task corresponds to the specified task format.
\end{itemize}

Upon completing the survey, the experts could expand their assessments by noting any additional irregularities or remarks. This section, intentionally left open-ended, was aimed at allowing the experts to share extra observations, recommendations, or issues they might have about the tasks presented.

\subsubsection{Parameter Selection}

To evaluate GQs systematically across multiple topics, we decided on generating a total of 30 questions across six distinct subjects within a university course ``\emph{Artificial Intelligence I}'', focusing on topics in \emph{symbolic AI} with otherwise fixed parameters in the prompt (see \Cref{sec:pipeline}).

\subparagraph{Difficulty and Cognitive Dimension} The majority of task generation systems documented in the literature focus solely on generating \emph{factual} questions that target the \texttt{remember} dimension \cite{kumarNovelFrameworkGeneration2023}. However, since our main objective in AGQ is to create learning materials capable of effectively enhancing our learner models, it is imperative to include tasks that assess at least the cognitive level of \texttt{understand}. Thus, we have opted to confine our investigation in this experiment to tasks that align with the \texttt{understand} dimension, assuming that the remaining dimensions are even more challenging to design questions for in our context (as for humans, so for LLMs). We fixed the \emph{difficulty level} to be \emph{``medium''} since we had already determined the model to largely generate relatively easy questions, which are consequently less informative when assessing students' mastery of a topic.

\subparagraph{Number of Questions} There is a trade-off regarding the number of questions generated \emph{per prompt} at once: The more questions generated, the more \emph{diverse} they can be expected to be. In contrast, generating fewer per prompt and prompting multiple times prevents immediate consideration of previously generated questions, making it difficult to minimize overlap between questions. However, the model often refuses to generate too many questions at once; it only replies with a few and instructs the user to prompt to yield more. The maximal number of GQs per prompt seems to depend on the topic or, presumably, the amount of course materials added to the prompt, but up to five questions at once seems to work consistently. We leave a more systematic investigation of this trade-off and the resulting quality of \emph{sets} of questions \emph{as a whole} to future work and fixed the number to be five GQs per topic (for a total of 30 GQs).

\subparagraph{Topics} We chose six topics from the course spread roughly uniformly across the table of contents of the course notes, with an attempt of making some ``broader'' and others more ``specific''; these being: 

\begin{enumerate} 
    \item Arc Consistency (constraint satisfaction problems), 
    \item Alpha-Beta Search (game play) 
    \item Semantics of Propositional Logic 
    \item Syntax of First-Order Logic 
    \item The STRIPS model (planning) 
    \item The \emph{delete} Relaxation (a heuristic in solving planning problems). 
\end{enumerate}

\section{Results and Discussion}
\label{sec:results}

\subsection{Question Quality}

The \textbf{question types} of the GQs were dominated by \emph{single choice} questions (12 out of 30), followed by \emph{multiple choice} questions (18 out of 30). The first consistently had three possible answers, while the latter almost always offered four options, usually with three distractors. The constant number of answers may be because the examples in the prompt also contain precisely this number of answers. With the MCQ, explicit care was taken to ensure that the example provided contained more than one correct answer. Nevertheless, the generated MCQs almost exclusively contained exactly three distractors. Notably, no fill-in-the-blank (FIB) questions appear in the generated questions. This is likely because designing meaningful FIB questions is significantly more challenging for the \texttt{understand} dimension, especially considering the necessary constraint that the answer be unambiguous.

The \textbf{FIT} of the GQs to the provided teaching material was evaluated by the experts as consistently given (28 out of 30), and most of them were assessed as \textbf{solvable} using the available teaching material (27 out of 30). In most cases, the GQs were evaluated as \textbf{clearly} and \textbf{unambiguously} formulated. Finally, most experts rated the GQs as relevant for achieving the learning objective.

However, the quality of the provided \textbf{feedback} was mixed. Frequently, no feedback was generated for an answer option. When it was, it was often not helpful in that it merely rephrased a wrong answer again -- i.e., the feedback to an incorrect choice $X$ was a variant of \emph{``No, it is not the case that $X$''}. This instance of uninformative feedback has also been observed and reported in other studies on LLM-feedback (e.g., \cite{lohrYouReNot2024}).

\subsection{Content Errors}

Eleven out of 30 GQs contained errors, mainly in the answer options and the feedback. These errors occurred particularly frequently in the topics \emph{arc consistency} and \emph{semantics of propositional logic}. We can only speculate on the reasons for this. Still, it seems likely that it correlates with the conceptual complexity of the topics or how well the topic is represented in the model's original training data -- semantics requires a much better understanding of the \emph{meaning} of logical statements \emph{and} the formal/mathematical mechanisms that \emph{provide} statements with meaning (\emph{interpretation functions}, \emph{models} and their formal definitions, etc), and there are multiple ways to formulate these mathematically. Similarly, to \emph{understand} arc consistency requires a solid intuition about how various constraints in a constraint satisfaction problem \emph{interact} with respect to the variables in a particular problem.

One crucial result to consider is that the model may produce superficially plausible questions that align with the ``aesthetic'' of a good question but are false in not necessarily trivial ways.

One particularly striking example is the following question on the topic of \emph{propositional logic} (generated in an earlier experiment),
\begin{mdframed}\itshape
    If we have already established $\neg B$, how can we use the Natural Deduction rule for implication ($\Rightarrow$) elimination to infer a new formula?

    {\large$\Box\mskip-11mu \lower0.1em\hbox{\itshape X}$} Given $B \Rightarrow C$, deduce $\neg C$
\end{mdframed}
where the answer above was considered the intended and only correct one.
This is the common fallacy of \emph{denying the antecedent}, but without understanding the semantics of propositional logic (or the natural deduction calculus) already, this is not immediately obvious.

Notable here is that the question is not only \emph{wrong}, but that it actively \emph{reinforces a common misconception} that may not necessarily be caught even by, e.g., student employees who assume they are sufficiently knowledgeable in the domain to evaluate questions, which is arguably even more dangerous than a merely \emph{wrong} question.

\subsection{Semantic Annotations}
We can distinguish between two kinds of annotations in \sTeX: \emph{Structural annotations} and \emph{relational annotations}. 

\textbf{Structural annotations}, such as \lstinline|\begin{sproblem}|, \lstinline|\begin{mcb}| (for MCQ-blocks),\\ \lstinline+\mcc[<T|F>,feedback={...}]{<answer>}+ (for an answer in a MCQ-block), \lstinline|\objective{<competency>}{<symbol>}| etc. should be present in all questions, or all questions of a particular question type, and are very schematic in the sense that they are not particular to a specific question, topic or concept (other than the one explicitly provided as a parameter to the prompt). These annotations seemingly pose no significant challenge to the model -- the few examples in the prompt are enough for the model to pick up on them immediately and use them consistently and correctly in (almost) all generated questions. Rarely does it hallucinate plausible but non-existent symbol names in the objectives.

This is \emph{not} the case for \textbf{relational annotations}, such as symbol or module references. To annotate a piece of text with a specific symbol, one has to know the module it is declared in, import it (using \lstinline|\usemodule{...}|), and then annotate the text using the \emph{name} of the symbol (e.g. ``\lstinline|$a$ \sr{addition}{plus} $b$|'' annotates the word \lstinline|plus| with the symbol named \emph{``addition''} which is currently in scope -- globally, there may be multiple symbols with the same name).
Naturally, how to do such relational annotations can not be directly inferred from a few \emph{generic} examples. Instead, humans \emph{and} AI models need a way to systematically look up symbols and their containing modules if we want to be able to refer to them in annotations.

As mentioned in \Cref{sec:stex}, we use these annotations to determine the prerequisites of learning objects, including questions, so ideally, we would want the model to annotate as much as possible. Our initial hope that the provided learning objects -- which should have ample occurrences of the relevant symbol names and module imports -- would be enough for the model to ``learn'' to use them correctly turned out to be clearly false.\footnote{
One reason for that could be the unfortunately inconsistent usage of annotations in the course materials in general, improvements of which happen frequently but are far from exhaustive.} Additional explicit instructions in the prompt also did not improve matters.

We therefore attempted to use the canonical way of giving models the ability to look up additional information: \emph{function calling} and \emph{retrieval-augmented generation}. OpenAI's API for GPT-4 offers explicit support for providing the model with a set of functions and determining which parameters they need. When prompted, the model will then either return an output as usual, \emph{or} a JSON object specifying (1) which function the model ``wants'' to call, and (2) the parameters the call should use. It is then up to the caller of the API to implement functionality that augments the \emph{following} prompt with the correct \emph{return value} for that function call.

We allowed the model to call a \texttt{search} function with a sequence of arbitrary strings as parameters. We augmented the prompt by adding a paragraph that instructed the model to use this function to determine how to semantically annotate any text that refers to a concept from the domain of the course. We then used existing functionality to search for any \emph{definitions} in our corpora and picked the top ten results for each of the string parameters. The search results were subsequently appended to the prompt in a new delimited section, the instruction to call the function was removed (to avoid potentially many subsequent calls or infinite loops), and the resulting prompt was fed back to the model.

This, too, unfortunately, did not have the desired effect; on the contrary. The quality of the questions generated got noticeably worse without improving the quality of the semantic annotations. We conjecture that by appending the search result and making the total prompt significantly longer, individual parts of the prompt have less influence on the output, e.g. the instructions at the beginning of the prompt. It is possible that more experimentation, possibly including repeating the instructions multiple times throughout the prompt, can lead to better results in the long run.\footnote{According to anecdotal ``LLM-folklore'', repeating important instructions in a long prompt is supposed to improve results.}

\subsection{Limitations}

We used a single model (GPT-4-turbo) to generate the questions for the evaluation. At the time of the study (March 2024), this model was -- in our opinion -- the most suitable model (see \Cref{sec:model-selection} for a detailed explanation). However, we explicitly do not rule out the possibility that there will be ongoing ``better'' models that are more suitable and can lead to better results. 
A further limitation of this study is that we did not involve students in evaluating the GQ. To obtain objective evidence of the quality of the tasks, a survey of students on the tasks would be helpful and is planned for the future. Especially when evaluating \emph{multiple choice} questions, for example, an assessment by (only) experts is not sufficient. Whether a distractor is good or bad usually only becomes apparent when the exercise is presented to students (who try to solve it). If learners never choose the distractor, it is a good indicator that it is too apparent and, therefore, unsuitable. At the time of evaluation, however, there was no possibility of recruiting suitable students from the course. Additionally, most generated questions were not ``ready to use'' on students without passing a ``human filter''. This means that we would no longer have evaluated the LLM's capabilities, but the experts' ability to filter the generated questions sensibly.

Furthermore, the study did not include any control items. As we did not aim to find out whether the experts were able to distinguish between LLM-generated and human-generated tasks the control items were of less importance. Their only function would have been to normalize the expert's rating, which was not necessary for our research questions.

\section{Conclusion}
\label{sec:conclusion}

In this experiment, we investigated the extent to which large language models are suitable for generating semantically annotated quiz questions for higher education from semantically annotated course materials that are (1) suitable for a specific course and (2) address \emph{understanding} a concept.

    Our research reveals significant limitations in the application of LLMs for this task. While generating questions that address \emph{remembering} \emph{factual} knowledge works well, creating questions that address \emph{understanding} remains a notable challenge -- which would need questions aiming for \emph{conceptual} knowledge, or asking for explanations rather than recall of a simple concept. Despite LLMs' ability to generate a range of questions, domain experts' need for extensive filtering underscores the models' inadequacies in autonomously generating educationally valuable content. The quality of questions generated by LLMs often does not meet educational standards, primarily due to the simplistic nature of the questions and the lack of valuable (and correct) feedback. Thus, the required careful review process by experts is both time-consuming and counter to the goal of automated content creation. In addition, errors in the content of the questions are challenging to detect for non-experts, as semantically incorrect output from LLMs is known to be very close to the ``truth'' \cite{sobieszekPlayingGamesAis2022}.

Our results show that generating \emph{structural}, semantic annotations works well. However, this success was not reflected in the case of \emph{relational} annotations, which exhibited poor integration, indicating a limitation in the LLM's ``ability'' to contextualize and link concepts within the generated content effectively.

The results of the present work make clear that while LLMs can contribute to the pool of learning materials, their current state requires significant human intervention to refine and validate the generated content. So, despite the promises systems using LLMs give, the human-in-the-loop remains crucial. Although the prompt was carefully designed and we provided a lot of static and dynamic context, the results still show weaknesses in quality and semantic annotation. Nevertheless, further studies with a high number of generated tasks and more experts might show that the automatic generation of questions following our approach could lead to more efficiency when used as an assistive system for the preparation process. 

\section*{Acknowledgements}
The work reported in this article was conducted as part of the VoLL-KI (see \url{https://voll-ki.de}) funded by the German Research/Education Ministry under grant 16DHBKI089. We also would like to thank Rakesh Kumar for the work in developing a significant portion of the front end for the question generator.
\printendnotes

\newpage

\bibliography{literature}

\end{document}